\documentclass[letterpaper]{article} % DO NOT CHANGE THIS
\usepackage{aaai2026}  % DO NOT CHANGE THIS
\usepackage{times}  % DO NOT CHANGE THIS
\usepackage{helvet}  % DO NOT CHANGE THIS
\usepackage{courier}  % DO NOT CHANGE THIS
\usepackage[hyphens]{url}  % DO NOT CHANGE THIS
\usepackage{graphicx} % DO NOT CHANGE THIS
\usepackage{natbib}  % DO NOT CHANGE THIS AND DO NOT ADD ANY OPTIONS TO IT
\usepackage{caption} % DO NOT CHANGE THIS AND DO NOT ADD ANY OPTIONS TO IT
\usepackage{algorithm}
\usepackage{algorithmic}
\usepackage{booktabs}
\usepackage{multirow}
\usepackage{colortbl}
\usepackage{xcolor}
\usepackage{verbatimbox}
\usepackage{amsmath}
\usepackage{newfloat}
\usepackage{listings}
\DeclareCaptionStyle{ruled}{labelfont=normalfont,labelsep=colon,strut=off} % DO NOT CHANGE THIS
\floatstyle{ruled}
\newfloat{listing}{tb}{lst}{}
\floatname{listing}{Listing}
\title{Building Domain-Specific Small Language Models via Guided Data Generation}
\author {
    Aman Kumar\textsuperscript{\rm 1},
    Ekant Muljibhai Amin\textsuperscript{\rm 2},
    Xian Yeow Lee\textsuperscript{\rm 1},
    Lasitha Vidyaratne\textsuperscript{\rm 1},
    Ahmed Farahat\textsuperscript{\rm 1},
    Dipanjan D Ghosh\textsuperscript{\rm 1},
    Yuta Koreeda\textsuperscript{\rm 2},
    Chetan Gupta\textsuperscript{\rm 1}
}
\affiliations {
    \textsuperscript{\rm 1}Hitachi America Ltd., Santa Clara, CA, USA\\
    \textsuperscript{\rm 2}Hitachi Ltd., Tokyo, Japan\\
}

\begin{document}

\maketitle

\begin{abstract}
Large Language Models (LLMs) have shown remarkable success in supporting a wide range of knowledge-intensive tasks. In specialized domains, there is growing interest in leveraging LLMs to assist subject matter experts with domain-specific challenges. However, deploying LLMs as SaaS solutions raises data privacy concerns, while many open-source models demand significant computational resources for effective domain adaptation and deployment. A promising alternative is to develop smaller, domain-specialized LLMs, though this approach is often constrained by the lack of high-quality domain-specific training data. In this work, we address these limitations by presenting a cost-efficient and scalable training pipeline that combines guided synthetic data generation from a small seed corpus with bottom-up domain data curation. Our pipeline integrates Domain-Adaptive Pretraining (DAPT), Domain-specific Supervised Fine-tuning (DSFT), and Direct Preference Optimization (DPO) to train effective small-scale models for specialized use cases. We demonstrate this approach through \textit{DiagnosticSLM}, a 3B-parameter domain-specific model tailored for fault diagnosis, root cause analysis, and repair recommendation in industrial settings. To evaluate model performance, we introduce four domain-specific benchmarks: multiple-choice questions (DiagnosticMCQ), question answering (DiagnosticQA), sentence completion (DiagnosticComp), and summarization (DiagnosticSum). \textit{DiagnosticSLM} achieves up to 25\% accuracy improvement over open-source models of comparable or larger size (2B-9B) on the MCQ task, while also outperforming or matching them in other tasks, demonstrating effective domain-specific reasoning and generalization capabilities.
\end{abstract}

% Uncomment the following to link to your code, datasets, an extended version or similar.
% You must keep this block between (not within) the abstract and the main body of the paper.
% \begin{links}
%     \link{Code}{https://aaai.org/example/code}
%     \link{Datasets}{https://aaai.org/example/datasets}
%     \link{Extended version}{https://aaai.org/example/extended-version}
% \end{links}
\begin{figure}[htbp]
  \centering
  \includegraphics[width=0.352\textwidth]{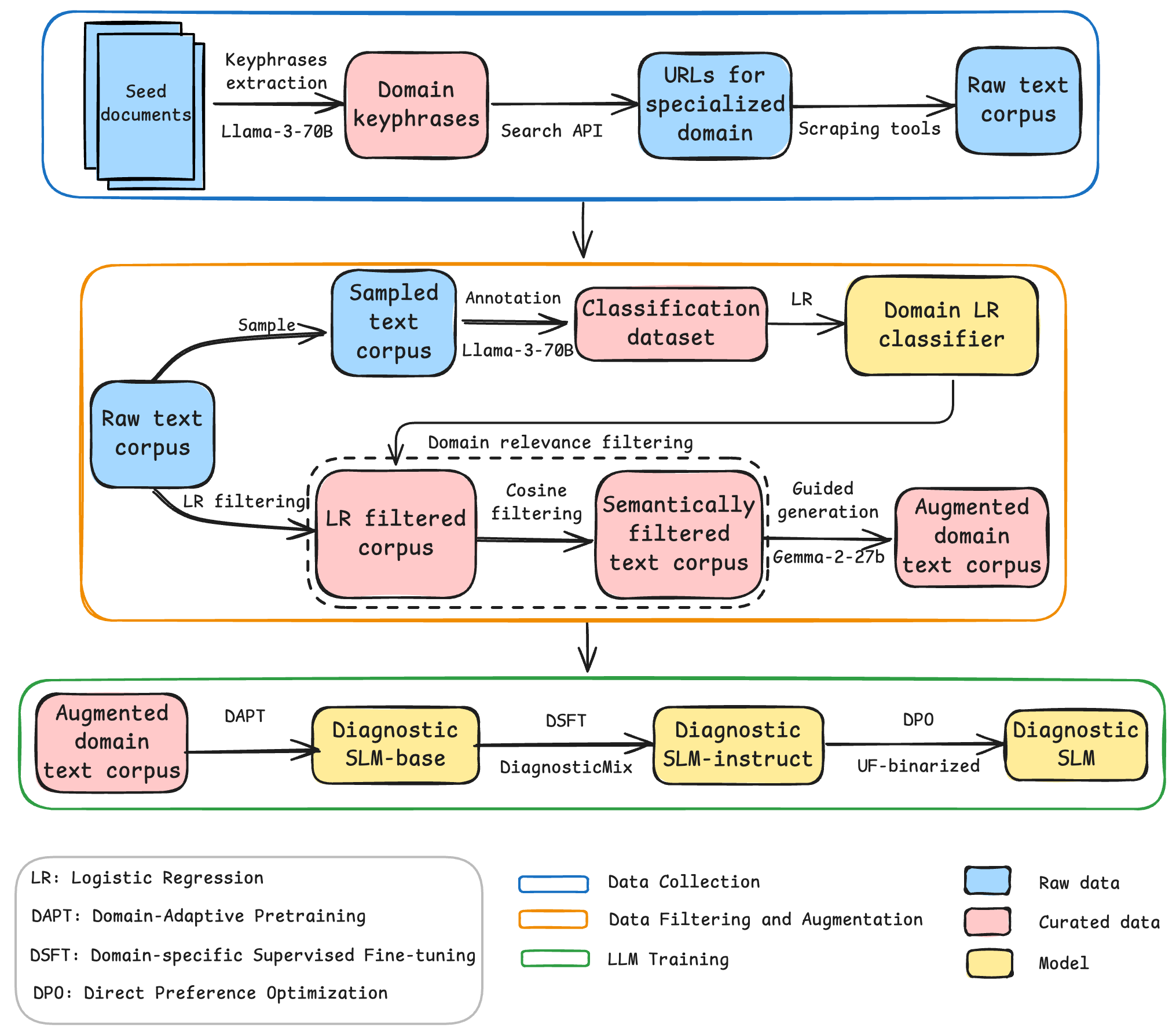}
  \caption{Overview of the \textit{DiagnosticSLM} pipeline. The figure illustrates the key stages of our approach, including domain-specific data collection, guided synthetic data generation using a teacher model, and a three-stage training process comprising Domain-Adaptive Pretraining (DAPT), Domain-specific Supervised Fine-tuning (DSFT), and Direct Preference Optimization (DPO).}

  \label{fig:intro}
\end{figure}

\section{Introduction}

Large Language Models (LLMs) have demonstrated remarkable potential in supporting a wide range of knowledge-intensive tasks across domains such as law~\cite{lai2024large}, finance~\cite{li2023large}, and medicine~\cite{thirunavukarasu2023large}. This success has prompted increasing interest in applying LLMs to specialized industrial domains, including fault diagnosis and repair. These domains are often hindered by fragmented documentation, shortage of skilled technicians, and the high cost and time investment required to train new personnel \cite{gao2015survey, nonaka1996knowledge}. While technicians performing low-skill operations may rely on manuals and standard procedures, high-skill tasks (such as fault identification, root cause analysis, and repair planning) require deep experiential knowledge that is rarely formalized and is often tacit \cite{argote2000knowledge}.

Despite the promise of LLMs, there are several barriers to their adoption in industrial settings. Proprietary models like GPT-4o \cite{hurst2024gpt}, Gemini \cite{team2023gemini}, and Claude \cite{bai2022constitutional} pose privacy risks due to cloud-only inference and lack of transparency, making them unsuitable for regulated environments. Open-source LLMs, though more accessible, often demand substantial computational resources for fine-tuning and deployment. Additionally, general-purpose models typically lack the domain-specific vocabulary, reasoning capability, and contextual grounding needed to support technicians in high-stakes diagnostic workflows~\cite{khodabandelou2024challenges}.

A promising alternative is to develop small, domain-specialized LLMs that can be deployed on-premise and customized for specific industrial use cases. Prior work on domain-adapted models such as LawGPT~\cite{nguyen2023brief}, BloombergGPT~\cite{wu2023bloomberggpt}, and ChipNeMo~\cite{liu2023chipnemo} has shown that targeted pretraining on specialized corpora significantly improves performance in expert domains. In particular, Domain-Adaptive Pretraining (DAPT)~\cite{gururangan2020don} followed by Supervised Fine-tuning (SFT) ~\cite{akatsuka2024rule} has emerged as a cost-effective strategy for domain adaptation.

However, a key bottleneck in this process is access to high-quality domain-specific training data. Industrial data is often siloed, proprietary, and inconsistently documented, limiting the ability to build or evaluate customized models. Moreover, curating clean, relevant subsets from massive corpora like Common Crawl is labor-intensive and prone to low recall \cite{remus2016domain}. Even with access to data, training from scratch or full fine-tuning of large models is often cost-prohibitive due to the extensive GPU hours and memory requirements involved \cite{xia2024understanding}.

To address these challenges, we propose a scalable pipeline for building domain-specialized small language models (SLMs).
% on a shoestring budget.\footnote{For reference, training Llama-3.2-3B from scratch requires an estimated 460,000 GPU hours, whereas our method consumed approximately 5,600 GPU hours, or about 1.2\% of that compute cost.} 
Our approach starts with bottom-up domain data curation and synthetic data generation from a curated seed corpus, followed by a three-stage training process to produce effective, lightweight models suitable for real-world deployment.

Our contributions are as follows:
(i) a pragmatic, low-resource pipeline that combines Domain-Adaptive Pretraining (DAPT), Domain-specific Supervised Fine-tuning (DSFT), and Direct Preference Optimization (DPO),  tailored for small on-premise LLMs deployed in regulated industrial domains such as automotive diagnostics;
(ii) a novel guided synthetic data augmentation strategy that leverages LLMs to enrich a bottom-up curated seed corpus for domain-specific data construction;
(iii) a targeted approach for generating supervised fine-tuning data aligned with key domain tasks; and
(iv) the introduction of four new domain-specific benchmarks, multiple-choice questions (DiagnosticMCQ), question answering (DiagnosticQA), sentence completion (DiagnosticComp), and summarization (DiagnosticSum), for evaluating LLMs in automotive diagnostics. An overview of our method is shown in Figure~\ref{fig:intro}.

\section{Methodology}
We adopted a three-step training pipeline to develop our proposed domain-specific SLM, \textit{DiagnosticSLM}. First, we perform additional pretraining of an open-source SLM, Llama-3.2-3B \cite{grattafiori2024llama,meta2024llama}, on a text corpus of Automotive domain. Next, we perform supervised fine-tuning using automotive domain related tasks adapted from general supervised fine-tuning Alpaca dataset \cite{alpaca}. Lastly, we perform direct preference optimization using the UltraFeedback dataset \cite{cui2023ultrafeedback}. For all experiments, we utilized $2\times$ NVIDIA RTX 4090 GPUs.

\subsection{Domain Adaptive Pretraining (DAPT)}
\label{sec:DAPT}

\subsubsection{Automotive Domain Data Curation}
\label{sec:ADDC}

We selected the automotive domain as the primary focus for diagnostics, with an emphasis on diagnostic procedures, repair operations, and core automotive concepts. We adopted a bottom-up data collection strategy, extracting keyphrases from internal documents and technical manuals using an internally hosted Llama-3-70B-Instruct model \cite{grattafiori2024llama} in a 4-bit configuration \cite{peft}. Outputs were verified by an expert for correctness and then used to guide targeted web searches via the Google Custom Search API. This process yielded approximately 403 million tokens from 706,971 webpages. Representative keyphrases are provided in the Appendix.

\subsubsection{Domain Relevance Filtering and Classification}
\label{sec:DRFC}
To further refine the dataset, we performed domain relevance filtering using a combination of LLM annotation and classical classification techniques. A random sample of 20,000 webpages was selected for annotation of the classification dataset. We used Llama-3-70B-Instruct model to label each instance as relevant or irrelevant to the automotive domain. The LLM annotated 11,621 samples as relevant and 8,379 as irrelevant, with the full annotation process taking approximately 22 GPU-hours. This highlights the considerable time and computational cost of large-scale LLM-based labeling, making it too expensive to annotate the entire dataset by experts. A manual review of a subset was conducted to validate label quality. While no corrections were applied, the low observed error rate and minimal review time (2 person-hours) supported continued use of the LLM-labeled data.

This labeled subset was split 80–20 into training and test sets. We trained a logistic regression classifier with L2 regularization ($C = 10$) using the SAGA solver. The resulting model achieved an accuracy of 88.2\% on the test set. Applying this classifier to the full dataset yielded 356,312 instances (50.39\%) classified as \textit{automotive-related}, and 350,624 (49.61\%) as \textit{non-automotive-related}.

Using eight topic-specific documents representing major automotive systems, we computed cosine similarity between each topic and all automotive-related samples, retaining those with a similarity score greater than 0.25. We also recovered possible false negatives by applying the same similarity check to the non-relevant set and retaining the top 20\%. After merging and deduplication, the curated dataset contained 387,572 samples (257M tokens). Details of the topic creation process and similarity computation are in Appendix.

\subsubsection{Guided Synthetic Data Augmentation}
\label{sec:GSDA}

To enhance the dataset, we employed a teacher model to expand the automotive content which is prompted to generate additional domain-relevant text and remove non-relevant portions. This augmentation step aimed to create a more comprehensive and detailed corpus, thereby improving the model’s ability to understand and generate automotive-specific information. A relatively smaller model, Gemma-2-27B \cite{team2024gemma}, was selected to facilitate efficient parallel inference across two GPUs. For data points exceeding the single-GPU context window, we partitioned the text into smaller chunks, processed them independently, and subsequently merged the outputs.

The augmentation process focused on enriching all relevant samples by prompting the teacher model to add factually accurate details and more comprehensive explanations, guided by its internal automotive knowledge. The underlying assumption is that the teacher model already possesses a certain amount of the necessary domain knowledge; rather than generating information from scratch, the prompts encourage the model to elaborate, clarify, and expand upon existing content. In this way, the teacher model plays a constructive role in enhancing the dataset by injecting technical depth and contextual breadth in a guided manner. This prompt-based strategy not only deepened the technical content and improved specificity across the dataset but also facilitated the filtering of non-relevant samples and removal of low-value, non-technical content inside the sample such as workshop addresses, geographic references, and irrelevant forum conversations. The entire expansion and modification pipeline consumed approximately 5,400 GPU-hours. Following augmentation, we applied fuzzy de-duplication using MinHash \cite{shrivastava2014defense} to remove redundant sentences including content from our automotive benchmark. The final automotive corpus comprised 206 million tokens.

\subsubsection{Model and framework}
\label{sec:MF}

We leveraged a 3B-parameter Llama-3.2 model for DAPT using our curated automotive corpus. The model was initialized from its publicly released pretrained weights. We employed the LlamaFactory Framework \cite{zheng2024llamafactory} for training and utilized fully sharded data parallelism (FSDP) \cite{zhao2023pytorch} via the Accelerate library \cite{accelerate} to distribute model parameters across GPUs.

\subsubsection{DAPT Training}
\label{sec:DAPTTrain}

We adopted a full-parameter fine-tuning strategy on the automotive domain-specific data. DAPT was conducted using the causal language modeling (CLM) objective, consistent with the pretraining objective of Llama-3 models. CLM can be defined as the negative log-likelihood of the next token given all previous tokens:

\begin{equation}
\mathcal{L}_{\text{CLM}} = - \sum_{t=1}^{T} \log P_\theta(x_t \mid x_{<t})
\end{equation}

where $x_t$ is the token at position $t$, $x_{<t}$ are all preceding tokens, and $P_\theta$ is the probability distribution parameterized by the model weights $\theta$.

Full-parameter fine-tuning was chosen over parameter-efficient methods (e.g., LoRA) to maximize domain alignment and capture deeper semantic shifts specific to the automotive domain, and was computationally feasible given the model’s relatively small size (3B parameters). The model was trained using the AdamW optimizer with a learning rate of $1 \times 10^{-4}$, and a cosine learning rate scheduler with 10\% warmup steps. We set the per-device batch size to 1 and used gradient accumulation over 8 steps, resulting in an effective global batch size of 16 across two GPUs. The input sequence length was set to 2,048 tokens, yielding a total of 16,384 tokens processed per forward pass. Training was conducted for 1 epoch, consisting of 5,789 steps in FP16 precision. The entire training process took approximately 118 GPU-hours. We call this model \textit{DiagnosticSLM-base}.

 \subsection{Domain-specific Supervised Fine-tuning (DSFT)}
\label{sec:DSFT}

\subsubsection{DSFT Data Generation}
\label{sec:DSFTDG}

To adapt the DAPT model to task-specific objectives, we constructed a domain-specific instruction-tuning dataset for the automotive domain. We curated 10 distinct task types and provided three example prompts per task to the GPT-4o model. These prompts guided the generation of instruction–response pairs across eight automotive topics introduced earlier in the data collection phase (e.g., Engine Repair, Brakes, etc.).

Using this setup, we generated approximately 20,000 examples spanning 10 automotive-domain instruction tasks inspired by Alpaca dataset tailored for the automotive domain. Details of each task are provided in the Appendix.

To balance domain-specific and general instruction-following capabilities, we combined our domain task dataset with the 52,000 samples from the Alpaca dataset. This produced a mixed training set for ablation studies evaluating the effects of different dataset combinations. The final DSFT corpus comprises approximately 72,000 instruction–response pairs, and we refer to it as \textit{DiagnosticMix}.

\begin{table*}[htbp]
\centering
\caption{Comparison of model variants on the \textit{DiagnosticMCQ} evaluation using 5-shot prompting. Columns A, B, C, and D indicate how many times each model selected each answer option across the full set of questions.}
\label{tab:ase_scores}
\begin{tabular}{
    l
    >{\centering\arraybackslash}p{1cm}
    >{\centering\arraybackslash}p{3cm}
    >{\centering\arraybackslash}p{2.75cm}
    >{\centering\arraybackslash}p{0.85cm}
    >{\centering\arraybackslash}p{0.85cm}
    >{\centering\arraybackslash}p{0.85cm}
    >{\centering\arraybackslash}p{0.85cm}
}
\toprule
\textbf{Model} & \textbf{Params} & \textbf{Training Type} & \textbf{Accuracy (Correct/Total)} & \textbf{A} & \textbf{B} & \textbf{C} & \textbf{D} \\
\midrule
\emph{Ground Truth (label dist.)} & -- & -- & -- & 240 & 246 & 205 & 185 \\
\midrule
\textbf{DiagnosticSLM (ours)} & \textbf{3B} & \textbf{DAPT+DSFT+DPO} & \textbf{0.4532 (397/876)} & \textbf{288} & \textbf{311} & \textbf{201} & \textbf{76} \\
Phi-3.5-mini-instruct & 3.8B & Original Instruct & 0.4384 (384/876) & 105 & 203 & 372 & 196 \\
Phi-4-mini-instruct & 3.8B & Original Instruct & 0.4098 (359/876) & 91 & 200 & 364 & 221 \\
Gemma-2-2B-it & 2B & Original Instruct & 0.3733 (327/876) & 265 & 83 & 317 & 211 \\
Llama-3.2-3B-Instruct & 3B & Original Instruct & 0.3653 (320/876) & 166 & 130 & 426 & 154 \\
Qwen2.5-3B-Instruct & 3B & Original Instruct & 0.3630 (318/876) & 115 & 154 & 352 & 223 \\
Gemma-2-2B & 2B & Base & 0.3368 (295/876) & 173 & 406 & 166 & 131 \\
Qwen2.5-3B & 3B & Base & 0.3288 (288/876) & 112 & 122 & 414 & 161 \\
Llama-3.2-3B & 3B & Base & 0.2705 (237/876) & 155 & 63 & 621 & 37 \\
DiagnosticSLM-base (ours) & 3B & DAPT & 0.2352 (206/876) & 4 & 3 & 868 & 1 \\
\midrule
\textbf{Llama-3.1-8B-Instruct} & \textbf{8B} & \textbf{Original Instruct} & \textbf{0.4692 (411/876)} & \textbf{136} & \textbf{220} & \textbf{305} & \textbf{215} \\
Gemma-2-9b-It & 9B & Original Instruct & 0.4521 (396/876) & 147 & 152 & 338 & 162 \\
Ministral-8B-Instruct-2410 & 8B & Original Instruct & 0.4281 (375/876) & 153 & 118 & 385 & 216 \\
Llama-3.1-Tulu-3-8B & 8B & Original Instruct & 0.4041 (354/876) & 140 & 241 & 129 & 366 \\
c4ai-command-r7b-12-2024 & 7B & Original Instruct & 0.3984 (349/876) & 216 & 241 & 331 & 87 \\
Llama-3.1-8B & 8B & Base & 0.3095 (247/876) & 171 & 110 & 402 & 46 \\
\bottomrule
\end{tabular}
\end{table*}

\subsubsection{DSFT Training}
\label{sec:DSFTTrain}

We fine-tuned the DAPT model on the DSFT dataset using an auto-regressive next-token prediction objective. 

\begin{equation}
\mathcal{L}_{\text{SFT}} = - \sum_{t=1}^{T} \log P_\theta(y_t \mid y_{<t}, x)
\end{equation}

where $x$ is the instruction prompt and $y = (y_1, \ldots, y_T)$ is the ground-truth response. The model is trained to generate $y$ conditioned on $x$.

Training was performed using the LlamaFactory framework with nearly identical hyperparameter settings as DAPT, except for a reduced learning rate of $1 \times 10^{-5}$. The per-device batch size was set to 2, with gradient accumulation over 8 steps, and an effective global batch size of 32.

The model was trained for 1 epoch, totaling 2,130 training steps, using precision FP16. The complete training process took approximately 47 GPU-hours. We call this model \textit{DiagnosticSLM-instruct}.

\subsection{Direct Preference Optimization (DPO)}
\label{sec:DPO}

While DAPT and DSFT enable the \textit{DiagnosticSLM-instruct} to internalize domain knowledge, they do not guarantee that the model’s outputs align with human preferences. To address this, we performed additional fine-tuning through preference alignment using the DPO~\cite{rafailov2023direct}. DPO is an alternative to Reinforcement Learning from Human Feedback (RLHF)~\cite{ouyang2022training} that directly optimizes a language model to prefer responses that align with human-annotated preferences, without requiring a separate reward model or reinforcement learning loop. Given a set of preference pairs \((x, y_{\text{pos}}, y_{\text{neg}})\), where \(x\) is the input and \(y_{\text{pos}}\) and \(y_{\text{neg}}\) are the preferred and less preferred responses respectively, DPO fine-tunes the model by minimizing the following loss:

\begin{equation}
\mathcal{L}(\theta)= -\log \sigma\left( \beta \left( \log \pi_\theta(y_{\text{pos}} \mid x)-\log \pi_\theta(y_{\text{neg}} \mid x) \right) \right)
\end{equation}

where \(\pi_\theta\) is the model's output distribution, \(\sigma(\cdot)\) is the sigmoid function, and \(\beta\) is a temperature parameter controlling the sharpness of preference. This encourages the model to increase the log-probability gap between preferred and non-preferred responses. Compared to RLHF, DPO offers a simpler and more stable training paradigm, while maintaining strong empirical performance. We adopt DPO in our alignment stage due to its computational efficiency, reduced implementation complexity, and its ability to effectively guide the model toward preferred behaviors.

\subsubsection{Dataset}
\label{sec:dpodata}

We used the UltraFeedback Binarized dataset \cite{cui2023ultrafeedback} as a general-purpose source of pairwise preferences for DPO and apply it to further fine-tune \textit{DiagnosticSLM-instruct}. A brief dataset summary appears in Appendix. Because UltraFeedback is not domain-specific, in future work we will construct an automotive preference dataset and repeat DPO with that domain data.

\subsubsection{DPO Fine-tuning}
\label{sec:DPOFT}

We fine-tune the Llama-3.2-3B–based \textit{DiagnosticSLM-instruct} model using DPO with LoRA, selected due to computational constraints from the additional memory overhead of preference-pair inputs and gradient storage. Training is conducted for 1 epoch with a per-device batch size of 1, a learning rate of $1 \times 10^{-5}$, and a cosine learning rate scheduler with a 10\% warm-up ratio. We use bfloat16 precision training and apply a sigmoid-based preference loss with $\beta = 0.1$. For LoRA, we set the rank to 16 and the $\alpha$ to 4. The input sequence length is capped at 2,048 tokens, and data preprocessing is parallelized across 8 workers. Evaluation is performed every 500 steps, with 10\% of the dataset reserved for validation. The entire DPO training process took approximately 144 GPU-hours. We refer to this final model as \textit{DiagnosticSLM}.

\section{Evaluation}

% \subsection{Evaluation}
The Automotive Service Excellence (ASE) certification exams are widely regarded as the standard for assessing technical proficiency in automotive diagnostics, repair, and maintenance across vehicle systems \cite{thompson2014perceived}. To evaluate domain-specific knowledge coverage and enable quantitative performance assessment, we constructed four benchmarks: multiple-choice questions (DiagnosticMCQ), question answering (DiagnosticQA), sentence completion (DiagnosticComp), and summarization (DiagnosticSum), inspired by the ASE exam format and subject areas. \textit{DiagnosticMCQ} was built from commercially licensed ASE-aligned practice materials, while the other benchmarks were derived from it through reformatting and GPT-4o–generated variations, followed by subject matter expert validation to ensure technical accuracy. We applied strict de-duplication against all training corpora to prevent overlap and contamination. For all experiments, we compare our model with Llama-3.2 \cite{meta2024llama}, Qwen2.5 \cite{yang2024qwen2}, Phi-3.5/Phi-4 \cite{abdin2024phi}, and Gemma-2 \cite{team2024gemma}.

\subsection{Multiple Choice Question Task}
\label{sec:mcq-task}

DiagnosticMCQ contains 876 multiple-choice questions across various automotive subtopics, each with four answer choices and one correct ground truth. Each evaluation prompt includes five example QA pairs followed by a test question to guide the model in understanding the structure and answer format. Model predictions are compared against the ground truth to compute accuracy. As shown in Table~\ref{tab:ase_scores}, our final model, \textit{DiagnosticSLM}, achieves an accuracy of 45.32\% (397/876), surpassing the Llama-3.2-3B-Instruct baseline (36.53\%) by a substantial margin. Despite its smaller size, \textit{DiagnosticSLM} also outperforms several larger models, including Ministral (8B, 42.81\%), Llama3.1-Tulu (8B, 40.41\%), and C4ai-command-r (7B, 39.84\%) and performs on par with Gemma-2 (9B, 45.21\%). It also exceeds the performance of similarly sized or slightly larger models such as Phi-4-mini-instruct (3.8B, 40.98\%), Qwen2.5-3B-Instruct (3B, 36.30\%), and Gemma-2 (2B, 37.33\%).

We observed that nearly all base versions of models exhibit skewed option selection when evaluated on \textit{DiagnosticMCQ}, often favoring a specific choice regardless of the question content. This trend shifts noticeably in their instruction-tuned variants. We observed a similar trend in our own model: after DAPT, the model predominantly selected option C. However, following DSFT and DPO, the answer distribution becomes more balanced. For comparison, the distribution of correct options in the ground truth is also included in the Table~\ref{tab:ase_scores}.

\subsection{Question-Answering Task}

LLMs often exhibit selection biases, i.e.,\ token bias towards label tokens (A/B/C/D)~\cite{pos-bias1,pos-bias2}, positional bias in answer ordering~\cite{bias}, and context priming from in‐context examples~\cite{prompt-bias1}. To introduce variation in evaluation and better assess the instruction-following capabilities of the models, we converted the \textit{DiagnosticMCQ} dataset into a question-answering (QA) format using GPT-4o, referred to as \textit{DiagnosticQA} benchmark. In this setting, the answer choices are embedded directly within the question in natural language, which reduces surface-level biases by requiring models to interpret the full input and generate an explicit answer rather than selecting a label. An example question is shown in Appendix Figure~\ref{fig:ase-mcq-data-sample}. We evaluate models using accuracy, comparing their generated answers against the ground truth.

\begin{table}[htbp]
\caption{Accuracy comparison of different models evaluated on the \textit{DiagnosticQA} dataset using 5-shot prompting.}
\label{tab:diagnosticqa_table}
\centering
\begin{tabular}{
    l
    c
    >{\centering\arraybackslash}p{2.5cm}
}
\toprule
\textbf{Model} & \textbf{Params} & \textbf{Accuracy} \\
\midrule
\textbf{DiagnosticSLM (ours)} & \textbf{3B} & \textbf{0.3831} \\
Phi-4-mini-instruct & 3.8B & 0.3628 \\
Qwen2.5-3B-Instruct & 3B & 0.3299 \\
Gemma-2-2B-it & 2B & 0.3169 \\
Phi-3.5-mini-instruct & 3.8B & 0.2342 \\
Llama-3.2-3B-Instruct & 3B & 0.2208 \\
\bottomrule
\end{tabular}
\end{table}

To assess performance in this free-form QA setting, we evaluated multiple models with similar parameter sizes. As shown in Table~\ref{tab:diagnosticqa_table}, all models experienced a drop in accuracy when shifting from MCQ to QA format. For example, \textit{DiagnosticSLM} decreased from 45.32\% to 38.31\%, \textit{Phi-4} from 40.98\% to 36.28\%, \textit{Qwen2.5} from 36.30\% to 32.99\%, \textit{Gemma-2} from 37.33\% to 31.69\%, \textit{Phi-3.5} from 43.84\% to 23.42\%, and \textit{Llama-3.2} from 36.53\% to 22.08\%. Despite the overall performance degradation observed in the QA setting, \textit{DiagnosticSLM} consistently outperformed all other models by a substantial margin.

\subsection{Sentence Completion Task}

\label{sec:completion-task}

% We created another benchmark \emph{DiagnosticComp}, a sentence completion task designed to isolate and evaluate LLMs' domain‐specific knowledge by mitigating the reasoning complexity and selection biases inherent in the MCQ task.  (see Section~\ref{sec:mcq-task}).  

To further assess domain-specific knowledge with reduced bias, we introduce the \textit{DiagnosticComp} benchmark. Each question in \textit{DiagnosticMCQ} is reformulated into four natural language sentences, each as an independent completion prompt, by appending one answer option to the shared question stem. The completions are generated using GPT-4o. Appendix Figure~\ref{fig:ase-mcq-data-sample} shows a sample prefix and its four completions.

\begin{table}[htbp]
\caption{Accuracy comparison of different models evaluated on the \textit{DiagnosticComp} dataset using log-likelihood-based scoring.}
\label{tab:completion_accuracy_table}
\centering
\begin{tabular}{
    l
    c
    >{\centering\arraybackslash}p{2.1cm}
}
\toprule
\textbf{Model} & \textbf{Params} & \textbf{Accuracy} \\
\midrule
\textbf{DiagnosticSLM (ours)} & \textbf{3B} & \textbf{0.5556} \\
Phi-4-mini-instruct & 3.8B & 0.5128 \\
Phi-3.5-mini-instruct & 3.8B & 0.5076 \\
Llama-3.2-3B-Instruct & 3B & 0.4786 \\
Gemma-2-2B-it & 2B & 0.4701 \\
Qwen2.5-3B-Instruct & 3B & 0.4188 \\
\bottomrule
\end{tabular}
\end{table}

% following the guidelines in Appendix~\ref{app:prompt-completion}. 

For each sentence prompt, we extract the model’s output logits at every generation step, apply a softmax over the vocabulary dimension to obtain token-level probabilities $P(x_i\mid x_{<i})$, and compute the \emph{aggregate log-likelihood}:

\begin{equation}
  \log P(x_{1:n}\mid\text{prefix}) = \sum_{i=1}^n \log P(x_i\mid x_{<i}).
\end{equation}

Since all four prompts share the same prefix, differences in total log-likelihoods reflect only the option texts. The option with the highest probability score is selected.

To focus on surface-level knowledge, we filter out questions requiring multi-step or implicit reasoning based on: (1) patterns such as reasoning chains or negations, and (2) option length. Longer options tend to accumulate lower log-likelihoods, so we restrict options to a maximum of four words and remove MCQs with longer options. After filtering, the final \textit{DiagnosticComp} set contains 117 completion questions. As shown in Table~\ref{tab:completion_accuracy_table}, \textit{DiagnosticSLM} outperforms other models, achieving 55.56\% accuracy.

\subsection{Summarization Task}

To evaluate models’ ability to summarize domain-specific technical content, we introduce the \textit{DiagnosticSum} task. Each input consists of a 3–6 line ground-truth explanation from the \textit{DiagnosticMCQ} dataset. Using GPT-4o, we generate concise two-line summaries of these explanations as reference outputs. Models are prompted to produce two-line summaries of the input, aiming to preserve key technical information in compact form.

We assess summarization quality using lexical and semantic similarity metrics, including ROUGE-1, ROUGE-2, ROUGE-L, BLEU, BERTScore (F1), and cosine similarity. ROUGE and BLEU are computed using the rouge-score and NLTK libraries, while BERTScore and cosine similarity are calculated with bert-score and sentence-transformers (all-MiniLM-L6-v2 \cite{wang2020minilm}). Given the creative nature of summarization, we run five independent trials per model with a temperature of 0.5 to capture variability. Aggregated results are reported in Appendix Table~\ref{tab:summarization_scores}. Phi-4-mini-instruct achieves the highest performance, while \textit{DiagnosticSLM} remains competitive with models of similar size. One possible reason for the relatively lower performance of \textit{DiagnosticSLM} is the limited number of domain-specific training examples related to the summarization task.

\subsection{Dataset Ablation Study}

\begin{table}[htbp]
\caption{Ablation study on different datasets evaluated on \textit{DiagnosticMCQ} using 5-shot prompting. 
\textit{Note:} SFT refers to fine-tuning on the Alpaca dataset; DSFT refers to fine-tuning on the \textit{DiagnosticMix} dataset.}
\label{tab:ase_scores2}
\centering
\begin{tabular}{
    >{\centering\arraybackslash}p{2.6cm}
    >{\centering\arraybackslash}p{3cm}
    >{\centering\arraybackslash}p{1.3cm}
}
\toprule
\textbf{Model Name} & \textbf{Training Type} & \textbf{Accuracy} \\
\midrule
Llama-3.2-3B-Instruct & Original Instruct & 36.53 \\
Llama-3.2-3B & Base + SFT & 30.71 \\
Llama-3.2-3B & Base + DAPT + SFT & 37.79 \\
Llama-3.2-3B & Base + DSFT & 38.24 \\
DiagnosticSLM-instruct (ours) & Base + DAPT + DSFT & 44.41 \\
\textbf{DiagnosticSLM (ours)} & \textbf{Base + DAPT + DSFT + DPO} & \textbf{45.32} \\
\bottomrule
\end{tabular}
\end{table}

To evaluate the individual and combined effects of DAPT, DSFT, and DPO, we conducted a series of ablation studies using the Llama-3.2-3B model. Our results, in  Table~\ref{tab:ase_scores2}, demonstrate that neither DAPT nor DSFT alone is sufficient to achieve optimal performance. Specifically, applying DAPT followed by SFT using Alpaca dataset yields a modest improvement in \textit{DiagnosticMCQ} accuracy (37.79) compared to the original instruction-tuned model (36.53), indicating that DAPT introduces some domain-awareness. Similarly, fine-tuning the base model directly with \textit{DiagnosticMix} data without prior DAPT results in a comparable \textit{DiagnosticMCQ} accuracy of 38.24, suggesting that task-specific supervision alone is also moderately effective. However, combining DAPT with \textit{DiagnosticMix} DSFT leads to a substantial performance boost, achieving a \textit{DiagnosticMCQ} accuracy of 44.41, demonstrating that domain pretraining and domain-specific supervision work synergistically. Further incorporating DPO pushes the \textit{DiagnosticMCQ} accuracy even higher to 45.32, underscoring the added value of preference alignment in improving model behavior.

\section{Pathway to Deployment}
\label{sec:path_to_deplyment}

\begin{figure}[htbp]
  \centering
  \includegraphics[width=0.46\textwidth]{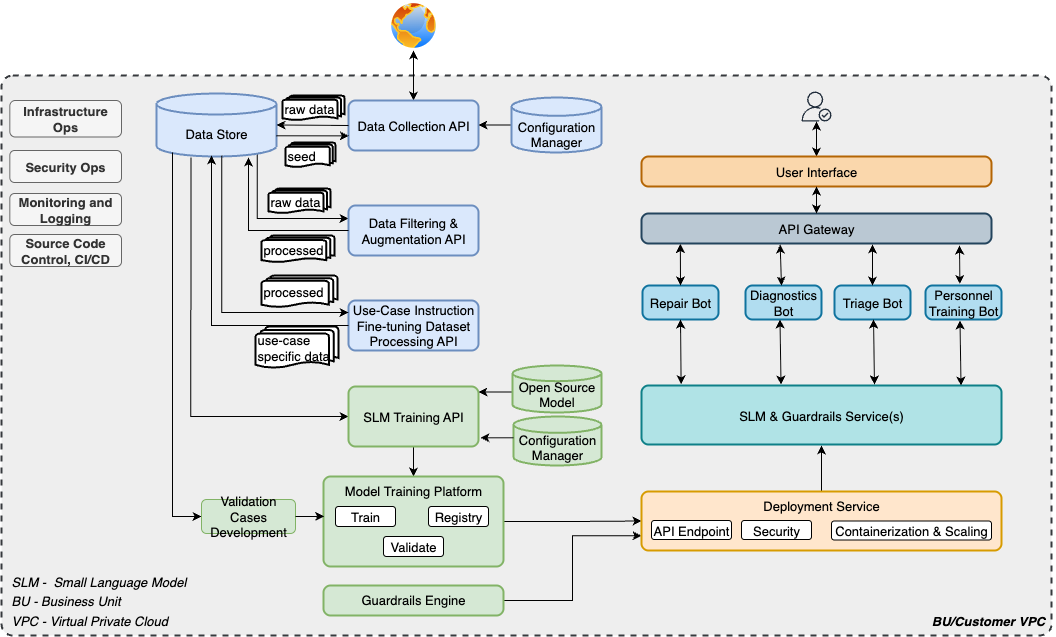}
  \caption{\textit{DiagnosticSLM} deployment architecture}

  \label{fig:architect}
\end{figure}

DiagnosticSLM is being prepared for on-premise deployment, and is designed to run inside customer or BU-managed infrastructure where data privacy, network isolation, and performance requirements must be met. Figure \ref{fig:architect} illustrates the end-to-end pipeline used to develop and operate the system. Data is collected from internal knowledge bases and selected external sources using the Data Collection API, then filtered and normalized through the Data Filtering and Augmentation APIs to create task-specific datasets. The SLM Training API and Model Training Platform manage model training, validation, and version-controlled registration, with a Guardrails Engine enforcing domain constraints, including abstention when model confidence is low.

For deployment, models are packaged as containerized inference services and hosted in customer or Business Unit (BU) virtual private cloud (VPC) environments using Kubernetes. The Deployment Service provisions API endpoints, authentication, authorization, and resource allocation. Inference primarily targets GPU nodes, with quantized CPU configurations available for lower-throughput environments. The deployed SLM and Guardrails services expose standard APIs consumed by downstream applications such as Repair, Diagnostics, Triage, and Personnel Training assistants through a unified API Gateway and User Interface. Monitoring of latency, abstention behavior, drift indicators, and error logs supports routine operation, while CI/CD workflows enable controlled updates and rollback. Post-deployment evaluation will examine how end-users interact with the system during routine maintenance and troubleshooting workflows, which may help identify where it reduces lookup effort or improves task efficiency.

\section{Conclusion}
\label{sec:conclusion}

We introduced \textit{DiagnosticSLM}, a small language model tailored for diagnostics and repair to assist frontline workers, and it is lightweight for edge deployment in industrial environments. Our three-stage training pipeline consisting of DAPT, DSFT, and DPO, enabled effective domain adaptation using curated and augmented data.  Across tasks such as multiple-choice questions, question-answering, sentence completion and summarization, \textit{DiagnosticSLM}outperformed or matched larger open-source models, underscoring the impact of specialized training on domain-specific corpora. Our ablation study highlights the complementary contributions of each training stage. While this work focused on validating the end-to-end pipeline, future work will incorporate retrieval-augmented inference and parameter-efficient fine-tuning techniques to further enhance performance and adaptability.

\bibliography{aaai2026}

\appendix

\section{Prompts}
\subsection{Prompts for DAPT Data Generation}

Prompt used for guided synthetic generation and knowledge distillation from the Gemma-2-27B-it teacher model.

\lstset{
  basicstyle=\ttfamily\footnotesize,
  breaklines=true,
  columns=fullflexible
}
\begin{lstlisting}
You are an automotive expert responsible for creating a high-quality automotive domain text corpus.  
This corpus must consist solely of technical content related to automotive topics.  
Carefully read the provided text. If any part of the text is unrelated to automotive topics (e.g., video games or non-automotive content), respond with ``NA''.  
Remove any irrelevant sentences from the text without explanation.  
For all relevant content, expand and improve the text by adding more factually accurate details and comprehensive explanations.  
Use your automotive knowledge to ensure the data expansion is correct and enhances the breadth and depth of the content.  
The goal is to augment the provided input to create high-quality, accurate, and detailed automotive content.
\end{lstlisting}

\subsection{Prompting Procedure for DSFT Data Generation}
\label{sec:dsft-prompting}

We generate a domain-specific instruction–response corpus by prompting a teacher model with three-shot, task-conditioned exemplars. We define a topic taxonomy of eight ASE-aligned areas (Engine Repair, Automatic Transmission/Transaxle, Manual Drive Train and Axles, Suspension \& Steering, Brakes, Electrical/Electronic Systems, Heating and Air Conditioning, Engine Performance) and a task set covering extractive QA, multiple-choice QA, question generation, free-form QA, true/false, sentence completion, sentiment, summarization, text generation, and topic classification. 

For each sample, we (i) uniformly select a topic and task, (ii) retrieve three representative in-context examples for the selected task, and (iii) call the teacher with the template below to produce one Alpaca-format triplet \texttt{\{instruction, input, output\}} conditioned on the chosen topic and task. We require outputs to remain within the automotive domain, avoid unsafe content, and follow the task’s output constraints. 

To ensure quality, we apply automatic checks: (1) domain filter (binary classifier) to reject off-domain generations, (2) length bounds for instruction and response, (3) format validation for task-specific constraints (e.g., MCQ label presence and exactly one correct choice), (4) lexical de-duplication with fuzzy matching, and (5) near-duplicate screening against training/evaluation splits. Accepted items are added to \textit{DiagnosticMix}.

\paragraph{Sampling logic.} Topic and task are sampled uniformly from the fixed sets; the three-shot exemplars are fixed per task to stabilize style. 

Prompt used for DSFT data generation:

\lstset{
  basicstyle=\ttfamily\footnotesize,
  breaklines=true,
  columns=fullflexible
}
\begin{lstlisting}
You are an expert automotive content generator. Produce ONE
instruction-tuning example in Alpaca format for the topic and task below.

Requirements:
- Stay strictly within the automotive domain for the specified TOPIC.
- Follow the TASK output rules exactly.
- Be factual, concise, and technically correct.
- Do not copy the examples verbatim. Create a new, distinct case.
- Return fields named: Instruction, Input, Response.

TOPIC: {TOPIC}
TASK:  {TASK}

Here are three in-context examples for this TASK:
### Instruction: {EX1_INSTR}
### Input: {EX1_INPUT}
### Response: {EX1_RESP}

### Instruction: {EX2_INSTR}
### Input: {EX2_INPUT}
### Response: {EX2_RESP}

### Instruction: {EX3_INSTR}
### Input: {EX3_INPUT}
### Response: {EX3_RESP}

Now generate a NEW example for the same TOPIC and TASK:

Return exactly:
Instruction: <one sentence task instruction>
Input: <short scenario or question>
Response: <the correct, task-compliant answer/output>
\end{lstlisting}

\section{Keyphrases for domain-specific web search}
\label{appendix:keyphrases}
\begin{itemize}
    \item Perform cylinder power balance tests
    \item Replace valve stem seals
    \item Cylinder Head and Valve Train Diagnosis and Repair
    \item Inspect cylinder deactivation system
    \item Check bearing preload to inspect, measure, and adjust
    \item Diagnose noise, vibration, and fluid leakage problems
\end{itemize}

\section{Automotive Tasks}
\label{appendix:auto_tasks}

We designed a set of diverse NLP tasks in the automotive context. The tasks are described below:

\begin{itemize}
    \item \textbf{Extractive Question Answering} – Identifying answers from a given automotive text.
    \item \textbf{Multiple-Choice Question Answering} – Selecting the correct answer from predefined options.
    \item \textbf{Question Generation} – Creating domain-specific questions based on a given text.
    \item \textbf{Open-Ended Question Answering} – Providing free-form responses to automotive queries.
    \item \textbf{True/False Classification} – Verifying the correctness of domain-specific statements.
    \item \textbf{Sentence Completion} – Predicting missing parts of automotive related text.
    \item \textbf{Sentiment Analysis and Summarization} – Detecting opinions in automotive discussions and summarizing content.
    \item \textbf{Text Generation} – Producing coherent, domain-specific automotive content.
    \item \textbf{Topic Classification} – Categorizing content into automotive subdomains (e.g., engine, transmission, brakes).
\end{itemize}

\begin{table*}[htbp]
\caption{Performance comparison of models on \textit{DiagnosticSum} benchmark dataset (mean ± std).}
\label{tab:summarization_scores}
\centering
\begin{tabular}{
    l
    >{\centering\arraybackslash}p{1.6cm}
    >{\centering\arraybackslash}p{1.6cm}
    >{\centering\arraybackslash}p{1.6cm}
    >{\centering\arraybackslash}p{1.6cm}
    >{\centering\arraybackslash}p{2.2cm}
    >{\centering\arraybackslash}p{1.8cm}
}
\toprule
\textbf{Model} & \textbf{Rouge-1} & \textbf{Rouge-2} & \textbf{Rouge-L} & \textbf{BLEU} & \textbf{BERTScoreF1} & \textbf{Cosine Sim} \\
\midrule
DiagnosticSLM (ours) & 0.5076 ± 0.0035 & 0.2691 ± 0.0024 & 0.3917 ± 0.0024 & 0.1815 ± 0.0029 & 0.9196 ± 0.0006 & 0.8522 ± 0.0019 \\
Gemma-2-2B-it & 0.5202 ± 0.0029 & 0.2654 ± 0.0028 & 0.3889 ± 0.0023 & 0.1938 ± 0.0030 & 0.9199 ± 0.0007 & 0.8583 ± 0.0021 \\
Llama-3.2-3B-Instruct & 0.4379 ± 0.0032 & 0.2310 ± 0.0024 & 0.3296 ± 0.0025 & 0.1167 ± 0.0019 & 0.8998 ± 0.0011 & 0.8229 ± 0.0019 \\
Phi-3.5-mini-instruct & 0.3796 ± 0.0018 & 0.2025 ± 0.0018 & 0.2796 ± 0.0015 & 0.1067 ± 0.0005 & 0.8902 ± 0.0020 & 0.8200 ± 0.0018 \\
\textbf{Phi-4-mini-instruct} & \textbf{0.5863 ± 0.0007} & \textbf{0.3455 ± 0.0024} & \textbf{0.4671 ± 0.0022} & \textbf{0.2757 ± 0.0031} & \textbf{0.9316 ± 0.0002} & \textbf{0.8882 ± 0.0013} \\
Qwen2.5-3B-Instruct & 0.4751 ± 0.0022 & 0.2448 ± 0.0017 & 0.3557 ± 0.0020 & 0.1386 ± 0.0015 & 0.9151 ± 0.0004 & 0.8802 ± 0.0010 \\
\bottomrule
\end{tabular}
\end{table*}

\section{Dataset Filtering}
\label{appendix:topic_filtering}
To enrich the automotive subset, using domain knowledge, we created eight topic-specific descriptive documents for:
\begin{enumerate}
    \item Engine Repair
    \item Automatic Transmission
    \item Manual Drive Train and Final Drive
    \item Suspension and Steering
    \item Brakes
    \item Automotive Electrical/Electronics
    \item Automotive Heating and Air Conditioning
    \item Engine Performance
\end{enumerate}

 \begin{figure}[htbp]
  \centering
  \includegraphics[width=0.46\textwidth]{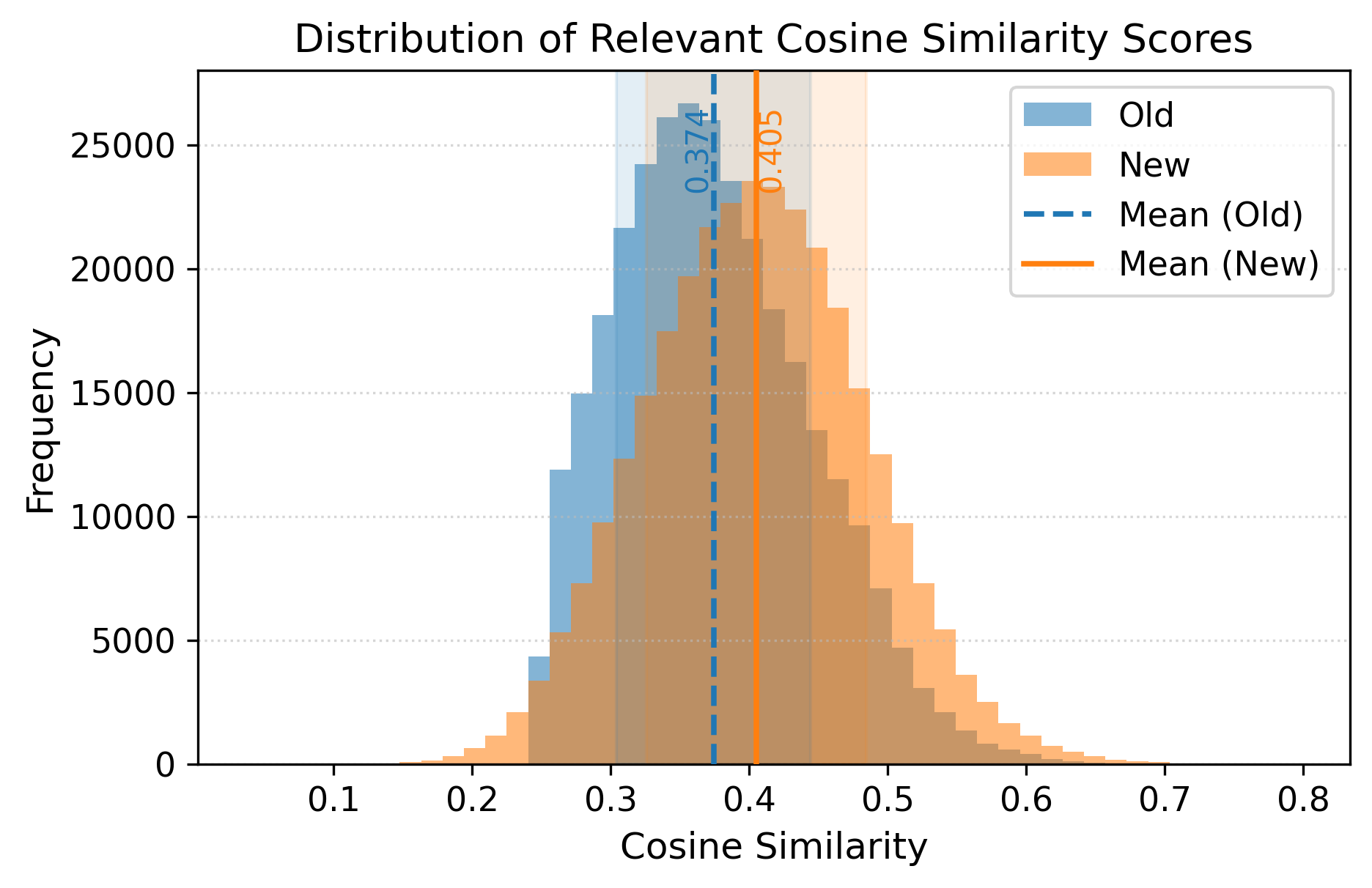}
  \caption{Shift of cosine similarity distribution from old to newly generated data}

  \label{fig:cosine_shift}
\end{figure}

  We computed the cosine similarity between the embeddings of each topic-specific document and the 356,312 previously classified automotive-related samples, yielding a total of 2,850,496 (8*356,312) similarity scores. We merged all scored instances, sorted them in decreasing order of similarity, and removed duplicate text entries to ensure uniqueness while preserving high relevance to at least one of the eight topics. The final set consists of unique samples, each associated with a corresponding similarity score. Samples with a cosine similarity greater than 0.25 were selected for inclusion in the main dataset. Figure \ref{fig:cosine_shift} shows the shift in distribution towards higher cosine similarity due to guided synthetic data generation and augmentation. About 66.46\% of the samples show increase in cosine similarity. 
 
 Moreover, to account for possible false negatives in the non-relevant set, we computed cosine similarity between the topic documents and all non-relevant samples. The top 20\% most similar entries were retained and merged with the previously filtered set. After merging and deduplication, the curated dataset contained 387,572 samples (257 million tokens).

\section{Base model selection}
To identify the most suitable base model for our pipeline, we initially experimented with three candidate models: Gemma-2B, Llama-3.2-1B, and Llama-3.2-3B. Each model underwent the same training pipeline consisting of DAPT and DSFT. We experimentally observed that both Gemma-2B and Llama-3.2-1B exhibited a decline in accuracy on \textit{DiagnosticMCQ} following DAPT and DSFT, indicating insufficient capacity to retain general language capabilities while learning domain-specific knowledge or are prone to overfitting and catastrophic forgetting during domain adaptation steps. In contrast, Llama-3.2-3B consistently demonstrated superior performance across the tasks. Based on these results, we selected Llama-3.2-3B as the foundation for all subsequent experiments. Having established Llama-3.2-3B as our base, we next analyzed how each training stage contributes to performance.

 \begin{figure}[htbp]
  \centering
  \includegraphics[width=0.45\textwidth]{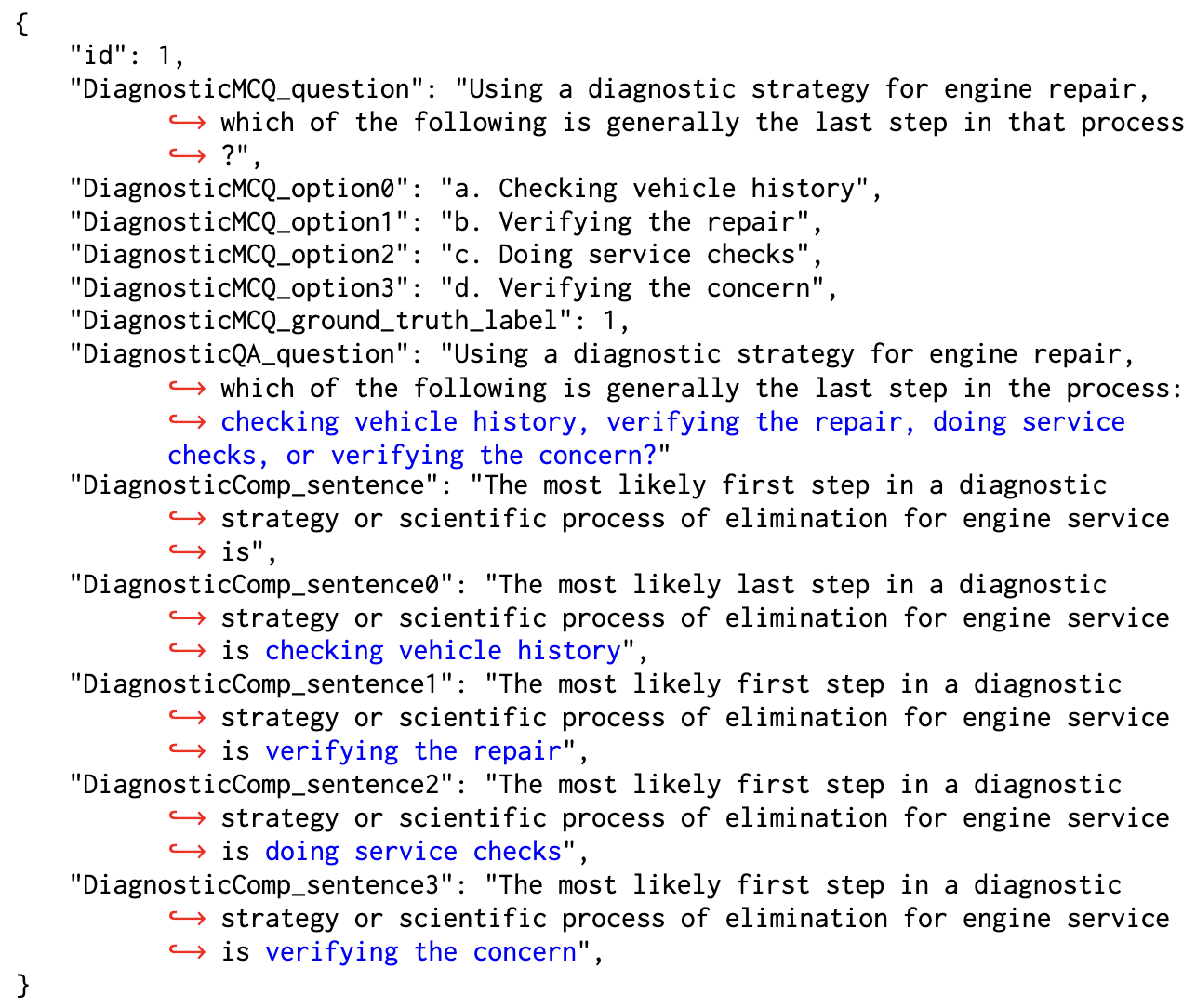}
  \caption{JSON representation of an example from our evaluation datasets.}

  \label{fig:ase-mcq-data-sample}
\end{figure}

\section{UltraFeedback Binarized dataset}

We utilize the UltraFeedback Binarized dataset \cite{cui2023ultrafeedback}, available on Hugging Face. The original UltraFeedback collection contains approximately 64{,}000 prompts, each with four completions generated by a mixture of proprietary and open-source language models. Completions were scored by GPT-4 on multiple criteria such as helpfulness and honesty. In the binarized release, the highest-scoring completion is labeled as the ``chosen'' response, and one of the remaining three is randomly selected as the ``rejected'' response, producing the pairwise preference format required for DPO. In this work, UltraFeedback serves as a general-purpose preference dataset; future work will construct an automotive preference dataset and re-run DPO with domain-specific preference pairs.

\end{document}